\documentclass[]{youtu} 

\usepackage{mathpazo}
\usepackage{graphicx}
\usepackage[numbers]{natbib}
\usepackage{times}
\usepackage[utf8]{inputenc} 
\usepackage[T1]{fontenc}    
\usepackage{url}            
\usepackage{booktabs}       
\usepackage{amsfonts}       
\usepackage{nicefrac}       
\usepackage{microtype}      
\usepackage{epsfig}
\usepackage{float}
\usepackage{multicol}
\usepackage{multirow}
\usepackage{amssymb}
\usepackage{amsmath}
\usepackage{wrapfig}
\usepackage{xspace}
\usepackage{color}
\usepackage{colortbl}
\usepackage[dvipsnames, svgnames, x11names, table]{xcolor}
\usepackage[misc]{ifsym}
\usepackage{makecell}
\usepackage{soul}
\usepackage{bm}
\usepackage{wrapfig}
\usepackage{subcaption, overpic, textpos}
\usepackage{paralist}
\usepackage{tabularx}

\usepackage{algorithm}
\usepackage{algorithmic}
\usepackage{adjustbox}
\usepackage{pifont}

\usepackage[capitalize]{cleveref}
\crefname{section}{Sec.}{Secs.}
\Crefname{section}{Section}{Sections}
\Crefname{table}{Table}{Tables}
\crefname{table}{Tab.}{Tabs.}

\newlength\savewidth

\renewcommand{\paragraph}[1]{\vspace{1.25mm}\noindent\textbf{#1}}

\definecolor{lightgray}{rgb}{0.8, 0.8, 0.8}
\definecolor{lgray}{rgb}{0.66, 0.66, 0.66}
\definecolor{whit_tab}{RGB}{255, 255, 255}
\definecolor{gray_tab}{RGB}{246, 246, 246}
\definecolor{oran_tab}{RGB}{252, 242, 237}
\definecolor{blue_tab}{RGB}{227, 240, 251}
\definecolor{lblu_tab}{RGB}{225, 235, 246}
\definecolor{orange_vitad}{RGB}{222, 131, 68}
\definecolor{blue_vitad}{RGB}{106, 153, 208}

\definecolor{trajectory_green}{RGB}{126, 171, 85}
\definecolor{trajectory_yellow}{RGB}{245, 194, 66}

\def\method{HumanVideo-MME}

\title{HumanVideo-MME: Benchmarking MLLMs for Human-Centric Video Understanding}

\author{\small Yuxuan Cai\textsuperscript{$\ast$}, Jiangning Zhang\textsuperscript{$\ast$}, Zhenye Gan, Qingdong He, Xiaobin Hu,  Junwei Zhu, Yabiao Wang, \\ Chengjie Wang, Zhucun Xue, Chaoyou Fu, Xinwei He, Xiang Bai}

\date{September 30, 2025}
\sourcecode{https://github.com/Fantasyele/HumanVideo-MME}
\project{https://fantasyele.github.io/projects/HumanVideo-MME/}

\begin{document}

\abstract{Multimodal Large Language Models (MLLMs) have demonstrated significant advances in visual understanding tasks involving both images and videos. However, their capacity to comprehend human-centric video data remains underexplored, primarily due to the absence of comprehensive and high-quality evaluation benchmarks. 
Existing human-centric benchmarks predominantly emphasize video generation quality and action recognition, while \textit{\textbf{overlooking essential perceptual and cognitive abilities}} required in human-centered scenarios. Furthermore, they are often limited by single-question paradigms and overly simplistic evaluation metrics. 
To address above limitations, we propose a modern \textbf{\method}, a rigorously curated benchmark designed to provide a more holistic evaluation of MLLMs in human-centric video understanding.
Compared to existing human-centric video benchmarks, our work offers the following key features: 
\textbf{(1) Diverse evaluation dimensions}: \method~encompasses 13 tasks, ranging from basic attribute perception (\textit{e.g.}, age estimation, emotion recognition) to advanced cognitive reasoning (e.g., social relationship prediction, intention prediction), enabling comprehensive assessment of model capabilities; 
\textbf{(2) Varied data types}: The benchmark includes multiple-choice, fill-in-blank, true/false, and open-ended question formats, combined with diverse evaluation metrics, to more accurately and robustly reflect model performance; 
\textbf{(3) Multi-domain video coverage}: The benchmark spans 50 distinct visual scenarios, enabling comprehensive evaluation across fine-grained scene variations; 
\textbf{(4) Temporal coverage}: The benchmark covers videos from short-term (10 seconds) to long-term (up to 30min) durations, supporting systematic analysis of models' temporal reasoning abilities across diverse contextual lengths.
We evaluate several advanced open-source MLLMs on the \method. While models excel in closed-form tasks, their performance drops sharply in open-ended generation, revealing a reliance on shallow patterns over genuine reasoning. In contrast, fill-in-blank and open-ended formats better expose reasoning challenges in human behavior understanding. By spanning diverse tasks and paradigms, \method~systematically reveals these limitations and facilitates the MLLM development.}

\maketitle
\vspace{-.1em}

\section{Introduction} \label{sec:introduction}

Recent advances in Multimodal Large Language Models (MLLMs)~\citep{hurst2024gpt,llava,chen2024internvl,qwen25vl, team2024gemini} have demonstrated remarkable capabilities in perceptual understanding and reasoning for general video comprehension tasks. Among various types of video data, human-centric videos represent a particularly critical domain due to their prevalence in real-world data. Compared to general video understanding, human-centric video understanding imposes greater challenges on models, as these tasks require not only the recognition of human actions and behaviors but also more sophisticated reasoning abilities. A systematic investigation of current MLLMs' capabilities and limitations in this domain is therefore critical for advancing both theoretical frameworks and practical applications.
However, existing benchmarks suffer from three fundamental limitations: 
\textbf{\textit{1)}} overly simplistic evaluation dimensions that fail to comprehensively cover the wide range of human-centric tasks; 
\textbf{\textit{2)}} restricted question-answer paradigms that overlook more complex and diverse reasoning needs; 
and \textbf{\textit{3)}} limited temporal coverage and scenario diversity, which hinder evaluation of MLLMs' generalization capabilities. These shortcomings inevitably undermine a comprehensive assessment of MLLMs' true potential in human-centric video understanding.

To bridge this gap, we propose {\method}, a human-centered video understanding evaluation benchmark. Compared to existing benchmarks~\citep{openhumanvid, humanvbench}, our benchmark introduces three key innovations: 
\textbf{\textit{1)}} \textbf{Diverse Evaluation Dimensions}: Covering 13 cognitive tasks, it systematically evaluates MLLMs' capabilities in both basic perception, such as age/gender recognition, emotion detection, action identification, and higher-order cognition, such as behavioral intention prediction and social relationship inference; 
\textbf{\textit{2)}} \textbf{Novel Evaluation Paradigms}: We propose a multi-paradigm framework that integrates various interaction modes-such as multiple-choice, fill-in-blank, and open-ended question answering—along with comprehensive quantitative metrics to provide holistic, application-oriented assessments; 
\textbf{\textit{3)}} \textbf{Spatiotemporal Coverage}: Spanning a wide range of 50 specidomains, with video durations ranging from 10 seconds to 30 minutes, this benchmark effectively evaluates models' ability to capture complex spatiotemporal relationships.

To evaluate the effectiveness of \method, we conduct extensive benchmarking across several state-of-the-art open-source MLLMs~\citep{chen2024internvl, qwen25vl, llavaonevision, cheng2024videollama}. Our results reveal that while several models achieve strong performance in Multiple-Choice (MC) and True/False (TF) formats, 
their performance notably degrades in Fill-In-Blank (FIB) and Open-Ended Questions (OEQ) formats.
For instance, while Qwen2.5-VL-32B~\citep{qwen25vl} achieves an impressive 94.77\% accuracy on multiple-choice causal reasoning tasks, its F1@1 score in open-ended causal generation drops close to zero (see \cref{tab:cot_results}). This stark contrast suggests that current MLLMs tend to rely heavily on superficial patterns or pretrained priors when solving closed-form questions, rather than engaging in genuine structural reasoning. 
Furthermore, the models exhibit consistently low accuracy on tasks requiring fine-grained visual perception, such as face recognition, underscoring persistent limitations in capturing subtle identity-related cues, which may be attribute to the scarcity of celebrity-focused data during pre-training.
These findings reveal two critical bottlenecks in the existing open-source MLLMs: weak generalization in generative tasks and insufficient grounding in fine-grained perception. By incorporating a diverse set of task types and question formats, \method~ systematically uncovers these limitations, and establishes a rigorous evaluation benchmark to guide the development of future MLLMs.

In summary, our contributions are as follow:
\begin{itemize}
  \item We construct {\method}, a large-scale benchmark tailored for human-centric video understanding. It covers 13 diverse tasks across perception and cognition, supports four QA paradigms (MC, FIB, TF, OEQ), and spans over more than 50 real-world scenarios, enabling spatiotemporal reasoning at multiple granularities.
  \item We introduce a novel composite evaluation metric for the causal reasoning task that integrates lexical accuracy, structural consistency, and LLMs-based semantic coherence scoring, enabling a more holistic and fine-grained assessment of generative reasoning capabilities.
  \item Extensive experiments reveal a stark contrast between performance in closed-form tasks and generative tasks. Our benchmark effectively exposes these limitations and guide future MLLMs toward better human-centric reasoning.
\end{itemize}

\section{Related Work} \label{sec:related_work}

\begin{figure}[tp]
    \centering
    \includegraphics[width=\linewidth]{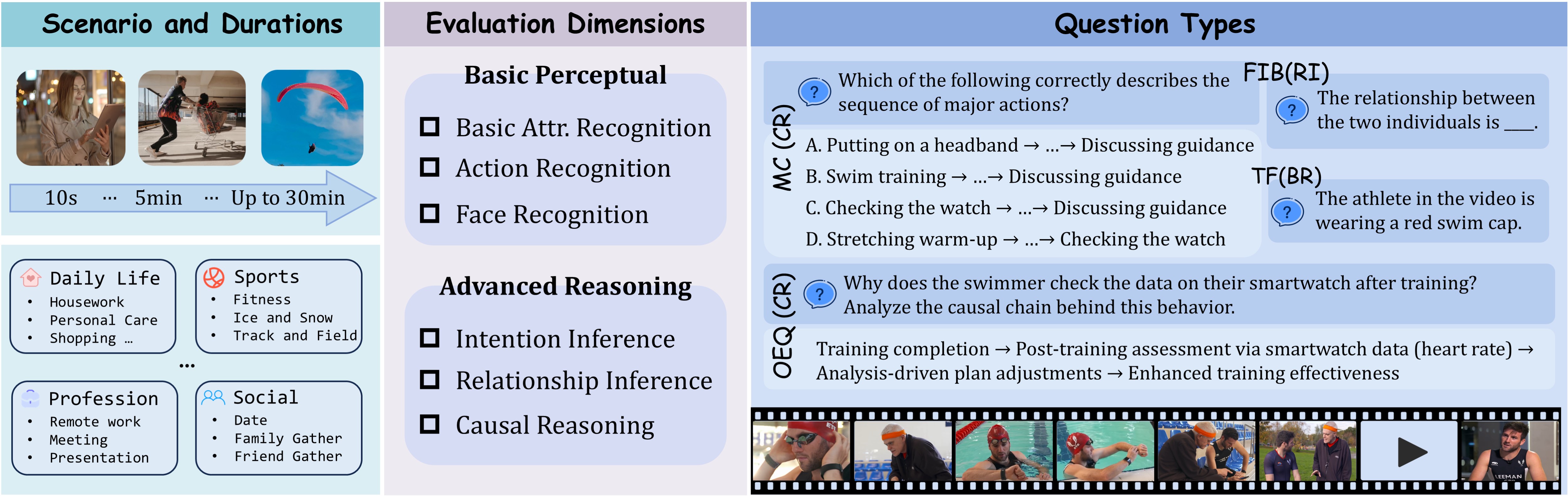}
    \caption{\textbf{Overview of \method} that spans diverse human-centric scenarios (50+ domains in 10s$\sim$30mins) and covers both basic perception and advanced reasoning tasks. It supports Multiple-Choice (MC), Fill-in-Blank (FIB), True/False (TF), and Open-Ended Questions (OEQ) to comprehensively evaluate MLLMs’ understanding and cognitive capabilities.}
    \label{fig:INTRO}
\end{figure}   

\subsection{Multimodal Large Language Models (MLLMs)}

The advent of large language models (LLMs)~\citep{llama, achiam2023gpt} has catalyzed significant breakthroughs in MLLMs~\citep{qwen25vl,chen2024internvl,hurst2024gpt,cai2024llava,llava15,team2024gemini,ye2024mplug,BLIP2} for visual-language understanding. In image-based tasks, MLLMs typically project visual features into the language space and leverage LLMs to fuse and reason over multimodal information. In contrast, video understanding poses greater challenges due to the intrinsic complexity of temporal modeling, demanding more sophisticated mechanisms to capture both spatial and temporal dependencies. 
Recent efforts have led to the development of several Video MLLMs~\citep{videochat,Videogpt,videollava,xu2024pllava} aiming to improve the understanding of temporal visual information.  Video-LLaVA~\citep{videollava} introduces a pre-alignment mechanism between image and video features for multimodal instruction tuning, achieving unified modeling across visual modalities. VideoChat~\citep{videochat} utilizes video foundation models to encode videos as embeddings, and feeds them directly into an LLM, enabling end-to-end video question answering.

Despite these advancements in architecture,  the potential of current video MLLMs for human-centric video understanding remains largely underexplored. To address this gap, we propose a dedicated benchmark focusing on human-centric video comprehension, which systematically evaluates model performance across both basic perceptual tasks and higher-order cognitive reasoning challenges.

\subsection{MLLM Benchmarks}

In the field of MLLMs, numerous benchmarks have been developed to evaluate models' capabilities in both perception and cognition. As image-based MLLMs have demonstrated impressive performance across various benchmarks~\citep{vqa2,vizwiz, gqa,sqa,mmb,mmmu,mme,pope}, research attention has increasingly shifted toward video understanding, an inherently more complex domain, thus driving the need for high-quality video benchmarks. 
Existing video datasets~\citep{videomme, mangalam2023egoschema,li2024mvbench,ning2023video} primarily focus on general video understanding tasks, such as object detection tasks. However, human-centric videos, which dominate real-world video content, present significantly greater challenges for multimodal understanding. Unlike general video tasks, understanding human-centric videos often requires comprehensive reasoning about human behaviors, intentions, emotional states, and social relationships.

Current human-centric video benchmarks can be categorized into two types: video generation evaluation~\citep{yu2019activitynet,openhumanvid,humanvid,huang2024vbench} and understanding assessment~\citep{humanvbench}. HumanVid~\citep{humanvid} combines real and high-quality synthetic videos with precisely annotated human poses to support research in portrait animation. OpenHumanVid~\citep{openhumanvid} integrates structured text, skeletal motion data, and speech to improve consistency and semantic alignment in video generation. In the domain of video understanding HumanVBench~\citep{humanvbench} has recently been proposed to focus on inner emotions and their external manifestations, introducing several fine-grained tasks such as emotion recognition. 
However, existing benchmarks suffer from oversimplified evaluation dimensions, restricted questioning paradigms, and limited temporal/scenario coverage. These shortcomings inevitably hinder a comprehensive assessment of the true potential of MLLMs in human-centric video understanding.

\section{\method~} \label{sec:method}
\begin{figure}[htp]
    \centering
    \includegraphics[width=1\linewidth]{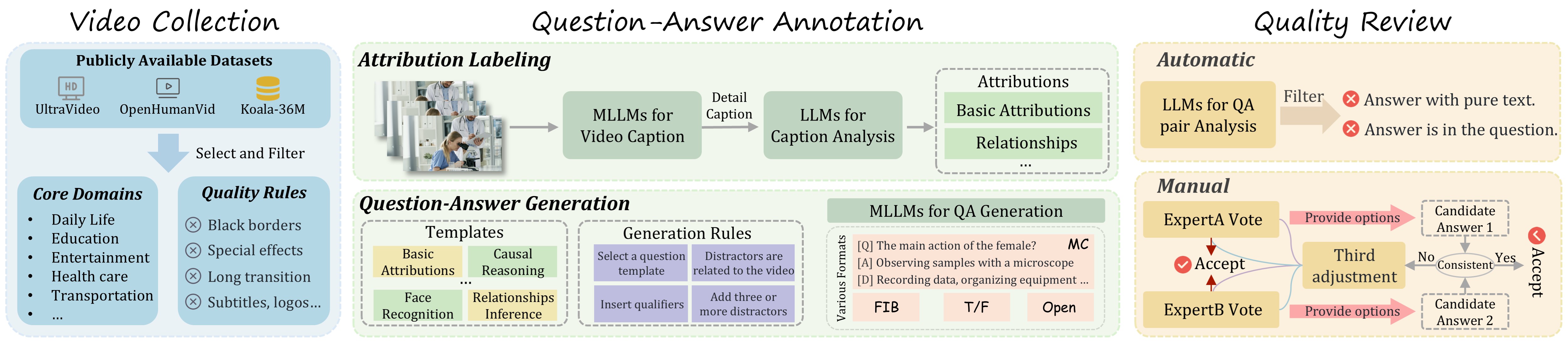}
    \caption{
    \textbf{\method~construction pipeline.} The benchmark is built through a three-stage pipeline: (1) large-scale Video Collection across diverse human-centric domains; (2) Automated QA annotation via MLLMs and structured templates; (3) a two-tier Quality Review combining automatic filtering and expert verification to ensure annotation reliability.
    }
    \label{fig:pipeline}
\end{figure}

{\method} dataset construction process involves three primary steps as shown in Fig.~\ref{fig:pipeline}: video collection, automated question-answer annotation, and quality review.

\subsection{Video Source and Collection}
\begin{figure}[htp]
    \centering
    \includegraphics[width=\linewidth]{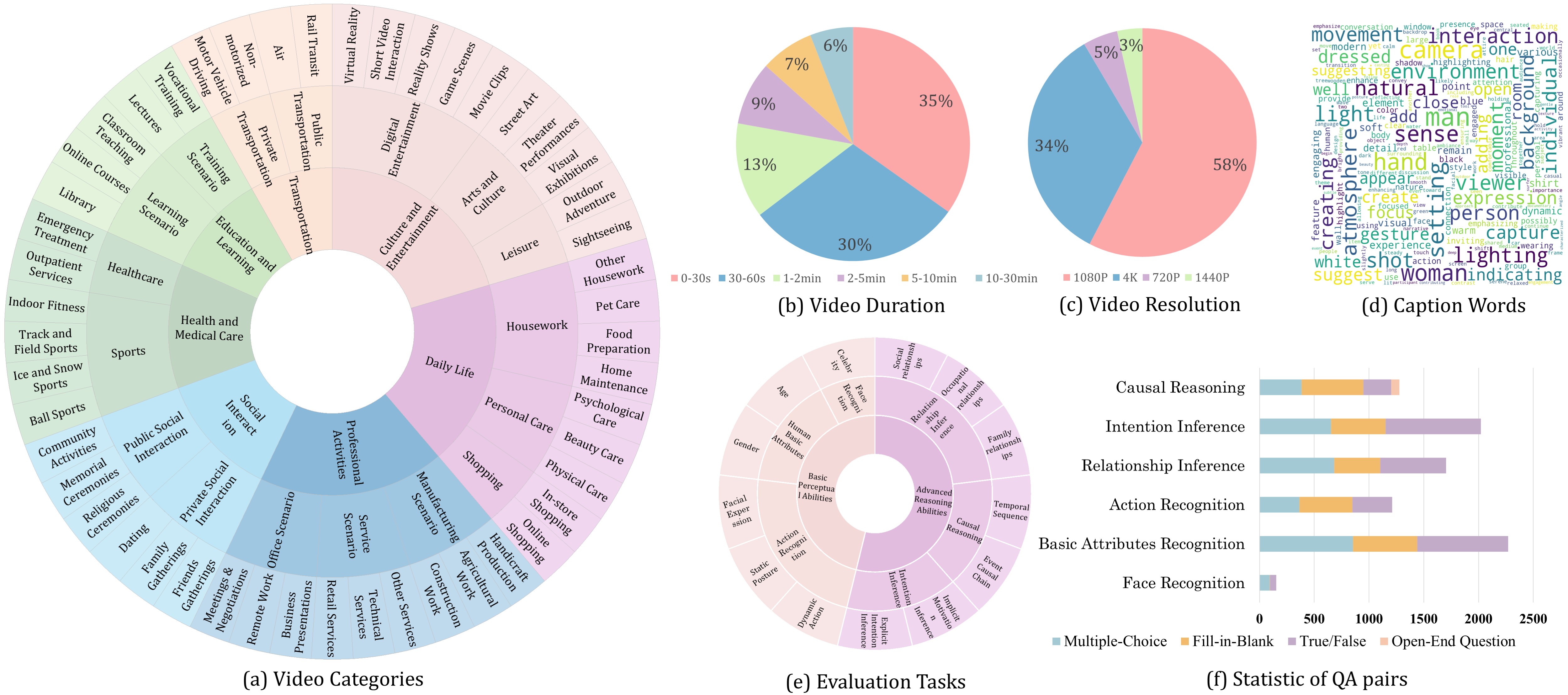}
    \caption{
    \textbf{Statistics of \method}. (a) Our benchmark covers 50+ human-centric categories across diverse domains. (b) Video durations range from short clips to long-form content. (c) Most videos are in high resolution (720P or above), supporting fine-grained visual analysis. (d) The caption vocabulary covers diverse semantic cues, emphasizing human appearance, actions, and interactions. (e) Our evaluation tasks span both perceptual and advanced reasoning abilities. (f) The statistic information of QA pairs. 
    }
    \label{fig:statistic}
\end{figure}

Our video data is sourced from publicly available datasets, including UltraVideo~\citep{xue2025ultravideo} and OpenHumanVid~\citep{openhumanvid}, as well as Koala-36M~\citep{koala36m}. To ensure that the dataset comprehensively covers a wide variety of human-centric contexts, we first define seven core domains to collect diverse data. These domains include: daily life, professional activities, social interactions, health and medical management, education and learning, transportation, and cultural entertainment. Each domain is further subdivided into specific scenarios to capture finer-grained contextual information. For instance, within the entertainment domain, we focus on scenes such as street performances. Ultimately, the \method~includes 50 meticulously categorized human-centered scenarios, and the full scene classification is shown in Fig.~\ref{fig:statistic}(a).

Sepcifically, data collection is conducted by 10 qualified researchers following strict quality control protocols. Videos were required to 
\textit{1)} align with predefined scenario classifications, 
\textit{2)} exclude artificial elements such as subtitles and black borders, and 
\textit{3)} maintain resolution above 720P. The final dataset comprises 1160 videos, with durations ranging from 10 seconds to 30 minutes, ensuring a diversity that spans from short-term interactions to longer-duration events.

\subsection{Automated Question-Answer Annotation Pipeline}

We meticulously design an automated annotation pipeline that consists of two key steps: 
\textit{1)} Attribute Labeling determines which types of evaluation tasks are suitable for constructing questions based on the video content. 
\textit{2)} Question-Answer Generation leverages the state-of-the-art MLLMs to automatically generate question-answer pairs for the relevant attributes, while the distractor generation further refines distractors, ensuring that they are not only distinct from the correct answers but also sufficiently challenging to enhance the evaluation's rigor.

\noindent\textbf{Video Captioning and Attribute Labeling.}
We categorize the evaluation dimensions of MLLMs into basic perception and advanced reasoning tasks. Basic perception tasks include human attribute recognition, face detection, and action recognition, while advanced reasoning tasks involve relationship inference, intent and motivation inference, and causal reasoning. To ensure that the generated questions are more targeted, we first perform Attribute Labeling on the video to identify the most relevant evaluation tasks.

To ensure high-quality attribute labeling grounded in both visual and semantic understanding, we adopt a two-stage model pipeline. Specifically, we employ the state-of-the-art Qwen2.5-VL-72B~\citep{qwen25vl}, to generate detailed video captions that encapsulate essential scene elements, human actions, and contextual cues from raw video inputs. These captions serve as a rich semantic representation of the visual content. Subsequently, we utilize the Qwen2.5-72B~\citep{qwen25}, which excels at language-based reasoning, to infer the most appropriate task-specific attribute labels based on the generated captions.

This decoupled framework which separates visual captioning from language-based label inference offers several key advantages. First, by embedding attribute labels within natural language descriptions, it enhances interpretability, making the allocation of evaluation dimensions more transparent and easier to understand. Second, it enables each model component to operate within its area of expertise, thereby mitigating error propagation: the MLLM focuses on extracting rich, structured semantic content from video, while the LLM performs high-level reasoning and task assignment based on textual cues. Overall, this approach bridges low-level perceptual signals and high-level reasoning objectives, enabling the generation of more coherent and targeted queries for complex, human-centric video understanding tasks.

\noindent\textbf{Question-Answer Generation.}
To ensure the diversity and accuracy of video question–answer pairs, we manually designed 5$\sim$10 varied question templates for each secondary evaluation task (\textit{e.g.}, action recognition in basic attribute perception, causal reasoning in advanced reasoning inference).
Prior to generating QA pairs, we first activate the corresponding attributes based on the annotations during the attribute labeling stage: only if a particular attribute is marked do we trigger the related question generation, thereby avoiding irrelevant or low-quality pairs.
During template-based question generation, we dynamically insert localization cues, such as ``the pedestrian on the far left'' or ``the man wearing a blue shirt'', according to scene complexity to eliminate ambiguity in multi-person contexts.
Next, we invoke the Qwen2.5-VL-72B~\citep{qwen25vl} to simultaneously produce the single correct answer and at least three distractors closely tied to key visual information; distractor design strategies include feature-detail variations (e.g., slight color or count adjustments) and temporal displacements (e.g., action-sequence perturbations) to heighten cognitive challenge.
Through this pipeline, we ensure that the generated QA pairs align tightly with the original video content while balancing evaluation depth and difficulty.

\subsection{Manual Quality Review and Adjustment}

To ensure the reliability and diversity of the dataset, we designed a rigorous review process consisting of following two main steps. 

\noindent\textbf{Automated Review.} Inspried by previous successes, we provide pure-text questions to the state-of-the-art LLMs like Qwen2.5~\citep{qwen25} and filter out low-quality questions where the answer is implicitly provided in the question or does not rely on understanding the video content. For example, questions like ``What is the gender of the male character wearing a blue T-shirt in the video?'' will not effectively evaluate the model's comprehension ability and are discarded in this step. The remaining questions are then subjected to further processing in the manual review phase.

\noindent\textbf{Manual Review.} 
We employ a two-stage independent human review combined with majority voting to ensure the quality of the dataset.
For MC and T/F questions, two experienced reviewers independently vote on each item, casting their votes and suggesting alternative options for any unreasonable choices.
When both reviewers reach consensus on the validity of the original answer or select the same alternative, the QA pair is accepted. In cases of disagreement, a third senior expert adjudicates the final decision. 
For fill-in-blank and open-ended questions, annotators are required to conduct a thorough assessment of the validity of all candidate answers. A question-answer pair is preserved only if both rounds of review confirm the reasonableness of all options. 
In summary, our review pipeline balances the efficiency of automated filtering with the rigor of manual inspection, substantially reducing labor costs while ensuring the accuracy of the question-answer pairs.

\subsection{Dataset Statistics and Analysis}

\noindent\textbf{Video Statistics.} 
Our benchmark dataset comprises 1,200 videos spanning multiple domains (\cref{fig:statistic}(a)), including social interactions, daily life scenarios, professional activities, sports, medical procedures, educational content, and entertainment. This comprehensive categorization captures the full spectrum of real-world human activities. 
Fig.~\ref{fig:statistic}(b) illustrates the temporal distribution of video durations, which range from brief clips (<30 seconds) to extensive segments (up to 30 minutes). This diversity in duration enables comprehensive evaluation of Video MLLMs across varying temporal dynamics, from instantaneous action recognition to complex sequence analysis requiring extended temporal comprehension.
Fig.~\ref{fig:statistic}(c) presents the distribution of video resolutions, which range from 720P to 4K and above. Notably, the majority of videos have resolutions exceeding 1080P, underscoring the high quality of our dataset.

\begin{table}[htp]
    \centering
    \caption{\textbf{Comparison of prominent benchmarks in the video domain.} We highlight key attributes, including domain scope (Open \textit{vs.} Human), the number of videos (\#Videos), the number of question-answer instances (\#QA Ins.), supported tasks, QA formats, evaluation metrics and resolution (Res). MC, FIB, TF, and OEQ are the abbreviation of multiple-choice, fill-in-the-blank, true/false, and open-ended questions, respectively.} 
    \label{tab:comparison}
    \renewcommand{\arraystretch}{1.1}
    \resizebox{1.0\linewidth}{!}{
\begin{tabular}{lcccccccc}
\toprule[0.17em]
Datasets     & Year    & Domain & \#Videos & \#QA Ins. & Tasks  & QA formats       & Metrics   & Res              \\
\midrule
Video-MME~\citep{videomme}     & 2024.05 & Open   & 900      & 2700          & Understanding & MC               & Acc   &        /       \\
OpenVid-1M~\citep{openvid1m}      & 2024.07 & Open   & 1M       & /             & Synthetic     & /                & /       &        /                   \\
Panda-70M~\citep{chen2024panda}        & 2024.02 & Open   & 70M      & /             & Synthetic     & /                & /           &        720P              \\
Koala-36M~\citep{koala36m}        & 2024 10 & Open   & 36M      & /             & Synthetic     & /                & /                &        720P        \\\hline
Tiktok-V4~\citep{chang2023magicdance}    & 2023.11 & Dance  & 350      & /             & /             & /                & /       &        /                 \\
OpenHumanVid~\citep{openhumanvid} & 2024.12 & Human  & 13.2M    & /             & Synthetic     & /                & /           &        720P             \\
HumanVBench~\citep{humanvbench}  & 2024.12 & Human  & /        & 2116          & Understanding & MC               & Acc.     &        /            \\
\rowcolor{blue_tab}
Ours         & 2025.05 & Human  & 1200     & 8700          & Understanding & MC, FIB, TF, OEQ  & Acc., F1, LLMs    &        720P-4K    \\ \hline
        \toprule[0.1em]
\end{tabular}
}
\end{table}

\noindent\textbf{QA Statistics.} 
Under our proposed \method, the designed task types encompass multiple QA formats, enabling comprehensive evaluation from basic perception to high-level cognition. Specifically, the Basic Attribute Recognition task is the largest, comprising 2,517 QA pairs across MC, FIB, and TF formats, reflecting broad coverage of fundamental human attribute perception. The Intention Inference and Relationship Inference tasks include 1,938 and 1,399 QA pairs, respectively—ranking second only to Basic Attribute Recognition—demonstrating our method’s strong coverage in advanced cognitive tasks. The Causal Reasoning task is the only evaluation involving open-ended questions, consisting of 70 OEQ samples (approximately 7.5\% of this task’s total), designed to evaluate the model’s high-level reasoning abilities beyond fixed-answer paradigms.
In addition, although Action Recognition and Face Recognition involve fewer samples, they serve as essential components at the perceptual level, contributing fine-grained understanding of human actions and identity, and thereby complementing the overall cognitive evaluation.

\noindent\textbf{Comparison with Related Benchmarks.} 
Table~\ref{tab:comparison} compares our proposed \method~with several existing benchmarks across different attributes, highlighting its unique advantages in human-centric multimodal video understanding. 
\textbf{\textit{1)} Comparison with open-domain benchmarks.} These benchmarks, such as OpenVid-1M~\citep{openvid1m} and Koala-36M~\citep{koala36m}, primarily focus on diverse video content from various domains. These datasets are large-scale, encompassing extensive video collections; for example, Panda contains over 70 million videos. Such benchmarks typically emphasize general video understanding tasks, including object detection and other perception-oriented objectives. However, they do not address the complexities inherent in human-centric video tasks, which require models to interpret human actions, emotions, intentions, and social interactions.
While these open-domain benchmarks are valuable for evaluating general video understanding, their lack of focus on human-centric aspects limits their ability to assess models in scenarios involving complex human behaviors and higher-order cognitive reasoning.

\textbf{\textit{2)} Comparison with Human-centric benchmarks.}
Unlike benchmarks such as HumanVid~\citep{humanvid} and OpenHumanVid~\citep{openhumanvid}, which primarily focus on evaluating video synthesis tasks, both HumanVBench~\citep{humanvbench} and our proposed \method~ are designed to assess video understanding tasks.   Compared to HumanVBench, \method~encompasses a broader range of cognitive tasks, including higher-order reasoning tasks such as relationship inference, intention prediction, and causal reasoning. This diversity makes \method~a more robust benchmark for evaluating human-centric video.  Another key distinction lies in temporal coverage: \method~accommodates video segments ranging from brief clips (approximately 10s) to extended sequences (up to 30 minutes), significantly surpassing HumanVBench's predominant focus on sub-10-second content. Furthermore, \method~employs a variety of question formats, including multiple choice (MC), fill-in-blank (FIB), true/false (TF), and open-ended questions (OEQ), as well as diverse evaluation metrics such as accuracy and LLM-based evaluation. This comprehensive approach enables a more thorough evaluation of model performance compared to benchmarks with a single question type or metric.
    
\section{Benchmarking MLLMs in {\method}} \label{sec:exp}
\subsection{Experimental Setting}

\subsubsection{Evaluation Details} 
\noindent\textbf{Evaluation MLLMs.} To comprehensively evaluate the understanding capabilities of open-source MLLMs in human-centric video scenarios, we benchmarked several state-of-the-art models, including Qwen2.5-VL series~\citep{qwen25vl}, InternVL2.5 series~\citep{chen2024internvl}, LLaVA-OneVision~\citep{llavaonevision}, LLaVA-Video~\citep{llavanext}, VideoLLaMA2~\citep{cheng2024videollama}.

\noindent\textbf{Implementation Details.} For MC and TF questions, we adopt the prompt: ``Answer with the option's letter from the given choices directly and only give the best option." This ensures that the model outputs a single, unambiguous letter corresponding to its top choice. For FIB tasks, we append the instruction ``Answer in short" to encourage concise and direct responses. For open-ended questions (OEQ), we guide the model with the prompt Use ``→'' to connect cause and effect explanations" to promote structured and interpretable outputs. All evaluations are conducted on 8 NVIDIA H20 GPUs. The batch size is set to 1.

\subsubsection{Metrics}
\noindent\textbf{MC and TF.} Following prior work in MLLMs, we adopt the standard Accuracy metric to evaluate model performance on MC and TF.

\noindent\textbf{FIB.}
To comprehensively evaluate the model’s ability to generate accurate and semantically appropriate answers in the cloze-style QA setting, we employ Top-1 evaluation strategies. Each question may correspond to multiple acceptable answers, and the ground-truth label for each sample is formulated as a set of valid candidate responses. The model is evaluated based on whether its prediction matches any of these candidates. 
Under the Top-1 setting, the model is required to produce a single answer per sample. We adopt the following standard metrics: 
\textit{Precision (Precision@1), Recall (Recall@1) and F1 (F1@1): } Precision@1 is defined as the proportion of predicted items that are correct, while Recall@1 measures the proportion of ground-truth answers that are covered. F1@1 is adopted to provide a balanced measure of both correctness and completeness.

\noindent\textbf{OEQ.}
To evaluate the causal reasoning capability of MLLMs, we adopt a hybrid evaluation framework that jointly considers factual accuracy, structural coherence, and semantic plausibility of generated causal chains. Each model prediction is compared against a human-annotated gold causal chain, typically represented as a sequence of events (e.g., ``A~$\rightarrow$~B~$\rightarrow$~C").

(1) Fuzzy Step-wise F1 Score ($Score_F$): we first compute a step-level F1 score that measures the overlap between predicted and ground-truth causal events using fuzzy token-based matching. This score accounts for lexical variation and reflects the factual correctness of the events comprising the chain.
(2) To evaluate structural consistency, we compute the Longest Common Subsequence (LCS) between the predicted ($P$) and reference chains ($G$). The final score is normalized by the length of the gold chain: $Score_O = \frac{\text{LCS}(P, G)}{|G|}$. 
(3) To evaluate the overall causal plausibility beyond step-level matching, we employ the state-of-the-art Qwen2.5-72B as a judge to rate the semantic coherence of each generated chain with respect to the reference. The model assigns a score ($Score_G$) from 0 to 5, where $Score_G = 0$ indicates no clear causal link between the predicted and reference chains, and $Score_G = 5$ represents  perfect semantic alignment and logically sound event sequencing.
We define a final composite score as a weighted sum of the three components, where $Score^{{\text{norm}}}_G$ is the normalized $Score_G$ ($Score^{{\text{norm}}}_G = \frac{Score_G}{5}$): 
\begin{equation}
Score = \alpha \cdot Score_F + \beta \cdot Score_O + \gamma \cdot Score^{{\text{norm}}}_G, 
\end{equation}
the default weights are $\alpha = 0.5$, $\beta = 0.3$, and $\gamma = 0.5$.

\subsection{Main Results}

\subsubsection{Evaluation under the MC and TF Question Formats}

Table~\ref{tab:mc_results} reports the performance of recent state-of-the-art MLLMs on {\method} under both Multiple-Choice (MC) and True/False (TF) formats. In the MC setting, models tend to perform well on high-level tasks such as Intention Inference (InI) and Causal Reasoning (CR), where Qwen2-VL-7B and Qwen2.5-VL-32B reaches the best 96.60\% / 95.66\% and 96.47\% / 94.77\% respectively. However, their performance significant drop in Face Recognition (FR) task  (42.83\% and 49.77\%), revealing their limitations in fine-grained facial understanding. At the same time, the scarcity of celebrity-centric samples in the pretraining corpus may also contribute to this degradation.
\textcolor{black}{This suggests that although large-scale MLLMs has been effectively equipped MLLMs with the ability to follow causal and relational logic, their capacity for fine-grained facial discrimination, particularly in capturing subtle identity-related cues under scarce celebrity samples, remains a major bottleneck.}
The performance trends in the True/False (TF) setting are generally consistent with those observed in MC. Compared to perceptual tasks, most models demonstrate better performance on advanced reasoning tasks.

\begin{table}[]
    \centering
    \caption{Performance of different open-sourced MLLMs on {\method} under the Multiple-Choice and True/False questions, respectively. BR, FR, AR are the abbreviate of Basic Attribute Recognition, Action Recognition, and Face Recognition. RI, InI, CR are the abbreviate of Relation Inference, Intention Inference, and Causal Reasoning, respectively.} 
    \label{tab:mc_results}
    \renewcommand{\arraystretch}{1.1}
    \resizebox{1.0\linewidth}{!}{
\begin{tabular}{lccccccc|ccccccc}
\toprule[0.17em]
\multicolumn{1}{c}{\multirow{2}{*}{Model}} & \multicolumn{7}{c}{Multiple-Choice}             & \multicolumn{7}{c}{True/False}    \\\cline{2-15}
\multicolumn{1}{c}{}                       & \multicolumn{1}{l}{BR} & \multicolumn{1}{c}{FR} & \multicolumn{1}{c}{AR} & \multicolumn{1}{c}{RI} & \multicolumn{1}{c}{InI} & \multicolumn{1}{c}{CR} & Avg                  & \multicolumn{1}{c}{BR} & \multicolumn{1}{c}{FR} & \multicolumn{1}{c}{AR} & \multicolumn{1}{c}{RI} & \multicolumn{1}{c}{InI} & \multicolumn{1}{c}{CR} & \multicolumn{1}{c}{Avg} \\
\midrule
VideoLLaMA2-7B       &      75.73 & 84.84 & 88.56 & 72.14 & 88.52 & 84.59 & 82.40 &    73.10 & 81.15 & 69.98 & 88.64 & 93.66 & 76.12 & 80.44       \\
LLaVAVideo-7B     &   87.85 & 65.59 & 97.46 & 95.86 & 94.73 & 91.13 & 88.77 &  71.90 & 75.94 & 75.23 & 90.41 & 92.30 & 82.18 & 81.33           \\
LLaVAOneVision-7B               &    87.01 & 85.30 & 95.61 & 95.38 & 95.45 & 92.07 & 91.80 &  73.68 & 84.90 & 79.05 & 91.54 & 95.60 & 83.33 & 84.68  \\
Qwen2-VL-7B        &  85.55 & 42.83 & 93.53 & 95.74 & 96.60 & 95.66 & 84.98  &  80.78 & 79.68 & 77.68 & 90.33 & 96.24 & 79.36 & 84.01       \\
Qwen2.5-VL-7B                              &   87.00 & 61.19 & 94.51 & 94.03 & 95.81 & 93.87 & 87.73 & 90.42 & 87.17 & 82.08 & 91.95 & 93.13 & 84.85 & 88.60           \\
Intern2.5-VL-8B      &     83.28 & 73.73 & 91.14 & 85.55 & 91.22 & 84.93 & 84.98 &   84.60 & 90.91 & 74.05 & 79.35 & 85.66 & 75.32 & 81.65           \\\hline
Qwen2.5-VL-32B      &  88.70 & 49.77 & 96.90 & 95.91 & 96.47 & 94.77 & 87.09 &  88.09 & 93.99 & 83.13 & 92.04 & 96.11 & 85.77 & 89.86 \\
\toprule[0.1em]
\end{tabular}
}
\end{table}

\subsubsection{Evaluation under the FIB Question Formats}

Table~\ref{tab:FiB_results} presents the performance of MLLMs on the {\method} under the Fill-in-Blank (FIB) question format.
In perception-level tasks, models generally achieve higher performance in basic attribute recognition (e.g., age, gender) compared to action recognition. For instance, Qwen2.5-VL-32B and LLaVAVideo-7B achieve Precision@1 scores of 61.2\% and 60.2\%, respectively, with corresponding F1@1 scores of 13.77\% and 13.73\%. These results highlight the strength of current MLLMs in low-level semantic understanding. However, in action recognition tasks, despite Qwen2.5-VL-32B achieving the highest Precision@1 (17.77\%), its F1@1 score remains low, indicating limited capacity in modeling dynamic behaviors.
In contrast, performance substantially degrades in higher-order reasoning tasks. For causal reasoning in particular, nearly all models exhibit F1@1 scores close to zero. The Intern2.5-VL series fails entirely on this task, while LLaVAVideo-7B performs best with an F1@1 of only 0.37. These results underscore the significant limitations of current MLLMs in modeling and generating coherent causal chains.

Notably, this trend contrasts with observations from the multiple-choice (MC) and true/false (TF) question formats. As shown in Table~\ref{tab:mc_results}, LLaVAOneVision-7B and Qwen2.5-VL-32B achieve MC accuracies of 92.07\% and 94.77\%, and TF accuracies of 83.33\% and 85.77\%, respectively, on the causal reasoning task. These findings suggest that in closed-form question settings, models can leverage language priors and pattern memorization from pretraining to match the correct answer, without necessarily performing genuine causal reasoning.

We hypothesize that under the MC or TF formats, models can often eliminate clearly incorrect choices or rely on surface-level keyword matching to select the correct answer. In contrast, the FIB format closely resembles natural language generation tasks, where models must produce responses without explicit options and rely solely on their internal reasoning capabilities. As such, FIB serves as a more challenging and discriminative benchmark for evaluating the true reasoning capacity of MLLMs.

\subsubsection{Evaluation under the OEQ Question Formats}

Table~\ref{tab:cot_results} presents the ability of various MLLMs to generate coherent causal chains in open-ended question settings. We observe substantial performance differences across models. In terms of $Score_F$, which quantifies the overlap between predicted and reference events via fuzzy token matching, most models perform comparably, ranging from 0.14 to 0.22, indicating that the lexical accuracy at the event level remains limited.
Similarly, the structural consistency score ($Score_O$) reveals that only the strongest models, Qwen2.5-VL-32B (0.51) and Qwen2.5-VL-7B (0.47), are able to retain a meaningful portion of the event sequence. The remaining models exhibit significantly weaker structural alignment.

Semantic plausibility ($Score_G$), assessed by the state-of-the-art LLM through the  metric, provides a clearer view of models’ higher-level reasoning capabilities. Qwen2.5-VL-32B achieves the highest $Score_G$ (0.69), followed by Qwen2.5-VL-7B (0.64) and Intern2.5-VL-8B (0.57), suggesting that some models are capable of generating causally plausible and semantically coherent chains even when token-level overlaps are limited.
In terms of overall performance, Qwen2.5-VL-32B leads with the highest average score (0.59), followed by Qwen2.5-VL-7B (0.57), while all other models score below 0.50.
Overall, the joint evaluation across $Score_F$, $Score_O$, and $Score_G$ provides a comprehensive assessment, capturing complementary aspects of model performance in causal reasoning.

\begin{table}[]
    \centering
    \caption{Performance of MLLMs on {\method} under the fill-in-blank questions. We employ Precision@1, Recall@1, and F1@1 for evaluation, respectively. } 
    \label{tab:FiB_results}
    \renewcommand{\arraystretch}{1.1}
    \resizebox{1.0\linewidth}{!}{
\begin{tabular}{lccccc}
\toprule[0.17em]
\multicolumn{1}{c}{\multirow{2}{*}{Model}} & \multicolumn{5}{c}{Tasks}            \\
\multicolumn{1}{l}{}                       & \multicolumn{1}{c}{Action Recognition} & \multicolumn{1}{c}{Basic Attribute Recognition} & \multicolumn{1}{c}{Causal Reasoning} & \multicolumn{1}{c}{Intention Inference} & \multicolumn{1}{c}{Relationship Inference} \\
\midrule
VideoLLaMA2-7B      & 4.40 / 0.50 / 0.90 & 10.73 / 1.47 / 2.53 & 0.37 / 0.03 / 0.07 & 0.13 / 0.00 / 0.03 & 6.27 / 0.73 / 1.37
                        \\
LLaVAVideo-7B                      & 10.57 / 1.37 / 2.33 & 60.23 / 7.90 / 13.73 & 1.93 / 0.20 / 0.37  & 1.53 / 0.13 / 0.30 & 13.37 / 1.87 / 3.13
                         \\
LLaVAOneVision-7B               & 4.27 / 0.50 / 0.87 & 49.37 / 6.50 / 11.30 & 0.00 / 0.00 / 0.00 & 0.77 / 0.07 / 0.13 & 3.77 / 0.43 / 0.83 \\
Qwen2-VL-7B                & 13.90 / 1.80 / 2.97 & 34.13 / 4.33 / 7.83 & 1.40 / 0.13 / 0.27  & 3.60 / 0.40 / 0.70 & 11.17 / 1.30 / 2.30 \\
Qwen2.5-VL-7B                             & 10.80 / 1.33 / 2.30 & 58.03 / 7.50 / 13.10 & 0.00 / 0.00 / 0.00 & 1.20 / 0.13 / 0.23 & 14.53 / 1.97 / 3.37                            \\
Intern2.5-VL-8B                           & 1.13 / 0.13 / 0.27 & 23.50 / 3.13 / 5.47 & 0.00 / 0.00 / 0.00 & 0.27 / 0.03 / 0.07 & 2.93 / 0.33 / 0.63 \\\hline
Qwen2.5-VL-32B                             & 17.17 / 2.00 / 3.50 & 61.17 / 7.87 / 13.77 & 0.00 / 0.00 / 0.00 & 2.67 / 0.27 / 0.50 & 15.87 / 1.87 / 3.37
                          \\
Intern2.5-VL-38B            & 3.13 / 0.37 / 0.63 & 28.60 / 3.57 / 6.23 & 0.00 / 0.00 / 0.00 & 0.00 / 0.00 / 0.00 & 5.63 / 0.90 / 1.50 \\\hline
\toprule[0.1em]
\end{tabular}
}
\end{table}

\begin{table}[]
    \centering
    \caption{Performance of MLLMs on {\method} under the open-ended questions for causal reasoning capacibility evaluation.} 
    \label{tab:cot_results}
    \renewcommand{\arraystretch}{1.1}
    \resizebox{1.0\linewidth}{!}{
\begin{tabular}{cccccccc}
\toprule[0.17em]
      & LLaVA-Video-7B & Qwen2-7B & Intern2.5-VL-8B & Intern2.5-VL-38B & VideoLLaMA2-7B & Qwen2.5-VL-7B & Qwen2.5-VL-32B \\
      \hline
$Score_F$      & 0.14           & 0.15     & 0.15            & 0.17             & 0.19           & 0.22          & 0.19           \\
$Score_O$     & 0.24           & 0.33     & 0.3             & 0.37             & 0.35           & 0.47          & 0.51           \\
$Score_G$     & 0.49           & 0.56     & 0.57            & 0.53             & 0.56           & 0.64          & 0.69           \\
$Score$ & 0.39           & 0.45     & 0.45            & 0.46             & 0.48           & 0.57          & 0.59           \\
\hline
\toprule[0.1em]
\end{tabular}
}
\end{table}

\section{Conclusion} \label{sec:conclusion}
In this paper, we presented {\method}, a benchmark encompassing 13 human-centric tasks—ranging from basic perception (e.g., age, emotion) to advanced reasoning (e.g., social relationship, intention, causal inference), and four QA formats (MC, FIB, TF, OEQ). By evaluating several state-of-the-art MLLMs, we revealed that existing MLLMs demonstrate relatively strong performance under MC and TF formats, but significantly underperform in generation-based FIB and OEQ tasks, especially in causal reasoning. This suggests that while current MLLMs may appear capable in structured tasks, they often rely on pattern heuristics instead of genuinely reasoning through the underlying causal or logical structure.

\textbf{Limitations and Border Impact}
We identify two major directions for future research. First, extending the benchmark to support comprehensive evaluation of generative models. Second, developing more fine-grained evaluation metrics, particularly for open-ended tasks—moving beyond accuracy to enable automatic assessment of reasoning coherence, factual grounding, and causal consistency. We believe these efforts are critical for advancing MLLMs toward deeper and more reliable understanding of complex human behaviors.

\section*{Contributions}

\paragraph{Authors~}
 Yuxuan Cai\textsuperscript{\rm 1$\ast$}, Jiangning Zhang\textsuperscript{\rm 2,3$\ast$}, Zhenye Gan\textsuperscript{\rm 3}, Qingdong He\textsuperscript{\rm 3}, Xiaobin Hu\textsuperscript{\rm 3},  Junwei Zhu\textsuperscript{\rm 3}, Yabiao Wang\textsuperscript{\rm 3}, \\ Chengjie Wang\textsuperscript{\rm 3}, Zhucun Xue\textsuperscript{\rm 2}, Chaoyou Fu\textsuperscript{\rm 4}, Xinwei He\textsuperscript{\rm 5}, Xiang Bai\textsuperscript{\rm 1}

\paragraph{Affiliations~}
\textsuperscript{\rm 1}Huazhong University of Science and Technology\quad \textsuperscript{\rm 2}Zhejiang University\quad \textsuperscript{\rm 3}Tencent Youtu Lab\quad \textsuperscript{\rm 4}Nanjing University\quad \textsuperscript{\rm 5}Huazhong Agricultural University

\setcitestyle{numbers,square}
\bibliography{youtu_bib}

\end{document}